\documentclass[conference,10pt]{IEEEtran}
\IEEEoverridecommandlockouts

\usepackage{cite}
\usepackage{amsmath,amssymb}
\usepackage{graphicx}
\usepackage{booktabs}
\usepackage{array}
\newcolumntype{L}[1]{>{\raggedright\arraybackslash}p{#1}}
\usepackage{url}
\usepackage[hidelinks]{hyperref}
\usepackage{xcolor}

\newcommand{\tbdate}{2.10}
\newcommand{\tbdvert}{59}

\let\oldthebibliography\thebibliography
\renewcommand{\thebibliography}[1]{%
  \oldthebibliography{#1}%
  \setlength{\itemsep}{0pt}%
  \setlength{\parsep}{0pt}%
  \setlength{\parskip}{0pt}}

\begin{document}

\title{MOLA LiDAR-Inertial Odometry (MOLA-LIO)\\ on the COMFORT Localization Benchmark}

\author{\IEEEauthorblockN{Jose Luis Blanco-Claraco}
\IEEEauthorblockA{\textit{Engineering Department} \\
\textit{University of Almer\'ia}, Spain \\
jlblanco@ual.es}
}

\maketitle

\begin{abstract}
This short report documents our entry to the COMFORT Localization Benchmark
(IROS 2026), evaluated on the GrandTour dataset \cite{frey2026grandtour}
recorded with the Boxi payload \cite{Tuna-Frey-Fu-RSS-25}. It extends MOLA-LO
\cite{blanco2025mola} into a LiDAR-inertial system that also ingests IMU and,
optionally, legged kinematic odometry. We describe the architecture, the streams
consumed, the local protocol that selected the submitted configuration, and the
measurements backing our real-time claim.
\end{abstract}


\section{Report Summary}

Table~\ref{tab:factsheet} answers, in one place, every item required by the
organizers.

\begin{table}[h]
\caption{Required declarations.}
\label{tab:factsheet}
\centering
\footnotesize
\begin{tabular}{@{}L{0.28\columnwidth}L{0.64\columnwidth}@{}}
\toprule
\textbf{Item} & \textbf{Answer} \\
\midrule
Method name        & MOLA-LIO \\
Affiliation type   & Academia (University of Almer\'ia) \\
Open / closed source & \textbf{Open source}, \url{https://github.com/MOLAorg/mola}; the GrandTour profile used here ships with \texttt{mola\_lidar\_odometry} \\
Framework          & MOLA modular SLAM framework (C++17, MRPT, \texttt{mp2p\_icp}, GTSAM) \\
Modalities used    & 3D LiDAR, IMU; \emph{no} legged odometry, \emph{no} camera, \emph{no} GNSS \\
Compute            & AMD Ryzen 7 2700X, 8 cores / 16 threads, 46\,GiB RAM \\
Real time          & \textbf{Yes}, all six missions: wall clock / duration 0.39--0.88, mean 0.60 (Sec.~\ref{sec:runtime}) \\
Post-processing    & None; the pipeline is causal and incremental \\
Loop closure       & \textbf{None}: no place recognition, revisit detection or pose-graph optimization \\
Novelty            & Published LO core \cite{blanco2025mola}; inertial fusion, cov-to-cov (GICP) matching and the asynchronous local-map rebuild are new \\
\bottomrule
\end{tabular}
\end{table}

\section{Method}

\subsection{Architecture}

MOLA declares a SLAM system as a graph of reusable blocks configured from YAML,
without writing code \cite{blanco2019mola,blanco2025mola}. Our entry uses three:
an \textbf{input} stage merging a mission's per-topic ROS\,1 bags into one
time-ordered stream; a \textbf{front end}, \texttt{mola\_lidar\_odometry},
registering each scan to the local map with \texttt{mp2p\_icp} using
generalized-ICP with point-wise covariances \cite{segal2009generalized}; and a
\textbf{state estimator} (\texttt{NavStateFilter}) supplying the ICP initial
guess, either a lightweight SE(3) filter (\emph{simple}) or a fixed-lag
IMU-preintegration smoother over a GTSAM iSAM2 factor graph \cite{gtsam}
(\emph{smoother}), which the submitted configuration uses over a 1.0\,s window.
The system runs online in a single pass: no offline re-optimization, batch
smoothing or map post-processing.

\subsection{From LO to LIO}

Three things differ from the LO system of \cite{blanco2025mola}.
\textbf{Inertial fusion}: the IMU levels the initial pose, constrains the
gravity direction, and in the smoother contributes preintegrated factors
between keyframes. \textbf{Cov-to-cov matching}: pairings are formed and
weighted from the local covariance of both clouds rather than from point
geometry alone. \textbf{Asynchronous local map}: the k-d tree rebalancing of
the local map is moved to a background thread, which cuts the time spent
inserting into the map by an order of magnitude (81.0\,s to 7.2\,s over a
436\,s mission) and removes the insertion spikes that would otherwise set the
per-scan worst case. Legged kinematic odometry was also implemented, as a full
SE(3) observation with loose velocity sigmas, but it did not improve the
held-out scores and is disabled here.

\section{Sensor Modalities and Topics}

Table~\ref{tab:topics} lists exactly which streams of the GrandTour bags are
consumed. Cameras and the GNSS/INS products (\texttt{cpt7\_ie\_*}) are
\emph{not} used.

\begin{table}[h]
\caption{Bags and topics consumed from each mission.}
\label{tab:topics}
\centering
\scriptsize
\begin{tabular}{@{}L{0.25\columnwidth}L{0.45\columnwidth}L{0.20\columnwidth}@{}}
\toprule
\textbf{Bag} & \textbf{Topic} & \textbf{Use} \\
\midrule
\texttt{*\_hesai\_undist} & \texttt{/boxi/hesai/}\newline\texttt{points\_undistorted} & LiDAR, 10\,Hz, motion-compensated \\
\texttt{*\_adis}          & \texttt{/boxi/adis/imu}                  & IMU, 200\,Hz \\
\texttt{*\_tf\_minimal}   & \texttt{/tf}, \texttt{/tf\_static}       & extrinsics \\
\bottomrule
\end{tabular}
\end{table}

The Hesai unit was selected after screening all three payload LiDARs on the
missions with a local reference, winning on every one; the ADIS16475 was
likewise preferred over the higher-rate STIM320. Scans come from the dataset's \texttt{points\_undistorted}
stream, already motion-compensated from the robot's legged kinematic-inertial
state estimate.

Poses are reported in the frame the benchmark requires: the estimate for the
body frame \texttt{base}, right-composed with the static
\texttt{base}\,$\rightarrow$\,\texttt{prism} transform and resampled to
200\,Hz. Resampling is an association device, not a source of information: under
the scorer's 10\,ms window a raw 10\,Hz file associates 868 of 4598 reference
samples on \texttt{heap-1} against 4576 once resampled, at a cost of 1.8\,mm
RMS.

\section{How the Submitted Configuration Was Chosen}

Because we started after the test-phase references became hidden, every configuration decision was
taken locally, on missions held out from the test set. Three reference classes
were used, in this order of trust: the \emph{total-station} prism trace of the
released missions, which is the class the benchmark itself scores against; a
\emph{GNSS/INS proxy}, used only where its own error is well below the effect
being screened; and \emph{reference-free consistency checks} otherwise, chiefly
attitude agreement against the robot's legged state estimator.

A single frozen configuration was uploaded for all six test missions, with \emph{no
per-mission tuning}. On the four held-out missions it reaches \tbdate\,cm ATE
against the prism reference, \tbdvert\% of which is vertical. That screen does
not separate the two state estimators (2.10 against 2.09 for the
lightweight filter); the smoother is submitted because it reports an honest
pose covariance where the filter reports a constant floor. Legged odometry, fused as an absolute pose, costs
1.1\,mm against a 0.1\,mm run-to-run floor, so it is off.
Table~\ref{tab:results} gives the official scores and what they cost.

\section{Runtime and Real-Time Operation}
\label{sec:runtime}

The submitted configuration is causal and single-pass (an additional tested configuration, reported nowhere here, applied an offline pose graph to these runs but was eventually dropped as a submission for this benchmark). Table~\ref{tab:results} reports the
ratio of wall-clock processing time to recorded mission duration on the
hardware of Table~\ref{tab:factsheet}, measured one run at a time on an idle
machine with the solver free to use all 16 hardware threads.

\begin{table}[h]
\caption{Per test mission: official score and what it cost. Ratio is wall clock over mission duration; CPU is in units of one core; latency is per scan, at a 100\,ms scan period.}
\label{tab:results}
\centering
\footnotesize
\begin{tabular}{@{}lcccccc@{}}
\toprule
& \textbf{ATE} & \textbf{Ratio} & \textbf{CPU} & \textbf{RSS} & \multicolumn{2}{c}{\textbf{Latency [ms]}} \\
\cmidrule(lr){6-7}
\textbf{Mission} & [cm] & & [cores] & [GB] & mean & worst \\
\midrule
\texttt{arc-2}  & 1.43 & 0.45 & 6.7 & 1.9 & 39 & 295 \\
\texttt{arc-7}  & 3.83 & 0.50 & 6.7 & 1.6 & 43 & 341 \\
\texttt{con-4}  & 9.43 & \textbf{0.88} & 7.6 & 4.3 & 82 & 544 \\
\texttt{eig-1}  & 3.32 & 0.66 & 7.0 & 2.3 & 61 & 308 \\
\texttt{snow-2} & 1.31 & 0.72 & 7.1 & 1.3 & 66 & 202 \\
\texttt{spx-2}  & 2.22 & 0.39 & 6.7 & 2.0 & 34 & 310 \\
\midrule
\textbf{Average} & \textbf{3.59} & \textbf{0.60} & 7.0 & 2.2 & 54 & 333 \\
\bottomrule
\end{tabular}
\end{table}

The margin is not uniform: \texttt{con-4}, whose radius-adaptive point budget
triples the cloud sizes, is the mission the claim turns on. Nor is the worst
case as comfortable as the mean: the slowest single scan takes 2 to 5.4 scan
periods and it is local-map insertion that sets it (452 of \texttt{con-4}'s
544\,ms), so those scans are absorbed as a short backlog, not dropped.
Registration dominates the average instead: ICP is 47--67\% of the per-scan
pipeline and 81--86\% of that forms and weights correspondences, against
5--11\% for the smoother and 2--6\% for map insertion. Reaching only 7.0 of 16
threads, the bound is serialization, not throughput.

\textbf{Reproducibility.} Repeating
the same configuration on the same machine moves the trajectory by 0.7 to
15.5\,mm RMS after alignment, because the local map's background rebuild thread
interleaves differently with registration. Forcing that rebuild synchronous
brings repeat agreement to about 1\,mm for roughly 30\% more wall clock. We
report the faster setting and this minor caveat rather than submitting a deterministic configuration that failed to achieve run-time for \texttt{con-4}.

\section{Discussion}

Two observations are worth recording for the benchmark's own sake.

\textbf{The ranked metric and our strongest metric disagree.} Our relative translation error, RTE=0.0098\,m, is
within 9\% of the best entry on the board, while our absolute error ATE is 2.4
times it. The post-alignment residual is a smooth, low-frequency deformation,
and it is mostly \emph{vertical}: against the total-station reference, 77--86\%
of the squared error lies in $z$ on the construction-site missions against
33--51\% elsewhere, and a reference-free diagnostic, the correction an offline
pose graph \emph{would} apply, orders the missions as our scores do. What is missing is an anchor on
verticality able to bound its drift over a ten-minute mission. This is clearly a future work to improve in our framework.

\textbf{The average over six missions is decided by its worst.} For us
\texttt{con-4} alone is 44\% of the average, and the spread across
near-identical configurations already exceeds the gap between leaderboard
positions. Per-mission ranks published alongside it would make the ranking far
less sensitive to a single hard sequence.

\scriptsize
\bibliographystyle{IEEEtran}
\bibliography{refs}

@article{frey2026grandtour,
  title   = {GrandTour: A Legged Robotics Dataset in the Wild for Multi-Modal Perception and State Estimation},
  author  = {Frey, Jonas and Tuna, Turcan and Fu, Frank and Patterson, Katharine and Xu, Tianao and Fallon, Maurice and Cadena, Cesar and Hutter, Marco},
  journal = {arXiv preprint arXiv:2602.18164},
  year    = {2026}
}

@inproceedings{Tuna-Frey-Fu-RSS-25,
  author    = {Jonas Frey and Turcan Tuna and Lanke Frank Tarimo Fu and Cedric Weibel and Katharine Patterson and Benjamin Krummenacher and Matthias M{\"u}ller and Julian Nubert and Maurice Fallon and Cesar Cadena and Marco Hutter},
  title     = {{Boxi: Design Decisions in the Context of Algorithmic Performance for Robotics}},
  booktitle = {Proceedings of Robotics: Science and Systems},
  year      = {2025},
  address   = {Los Angeles, United States},
  month     = {June}
}

@article{blanco2025mola,
  author  = {Blanco-Claraco, Jose Luis},
  title   = {A flexible framework for accurate {LiDAR} odometry, map manipulation, and localization},
  journal = {The International Journal of Robotics Research},
  year    = {2025},
  note    = {Also available as arXiv:2407.20465}
}

@inproceedings{blanco2019mola,
  author    = {Blanco-Claraco, Jose Luis},
  title     = {A Modular Optimization Framework for Localization and Mapping},
  booktitle = {Robotics: Science and Systems (RSS)},
  year      = {2019}
}

@misc{gtsam,
  author       = {Frank Dellaert and {GTSAM Contributors}},
  title        = {{borglab/gtsam}},
  howpublished = {\url{https://github.com/borglab/gtsam}},
  year         = {2022}
}

@article{segal2009generalized,
  title   = {Generalized-{ICP}},
  author  = {Segal, Aleksandr and Haehnel, Dirk and Thrun, Sebastian},
  journal = {Robotics: Science and Systems},
  year    = {2009}
}

\end{document}